# Hybrid Influence Diagrams Using Mixtures of Truncated Exponentials

**Barry R. Cobb and Prakash P. Shenoy**
University of Kansas School of Business
1300 Sunnyside Ave., Summerfield Hall
Lawrence, KS 66045-7585
*brcobb@ku.edu, pshenoy@ku.edu*

## Abstract

Mixtures of truncated exponentials (MTE) potentials are an alternative to discretization for representing continuous chance variables in influence diagrams. Also, MTE potentials can be used to approximate utility functions. This paper introduces MTE influence diagrams, which can represent decision problems without restrictions on the relationships between continuous and discrete chance variables, without limitations on the distributions of continuous chance variables, and without limitations on the nature of the utility functions. In MTE influence diagrams, all probability distributions and the joint utility function (or its multiplicative factors) are represented by MTE potentials and decision nodes are assumed to have discrete state spaces. MTE influence diagrams are solved by variable elimination using a fusion algorithm.

## 1 INTRODUCTION

An influence diagram is a compact graphical representation for a decision problem under uncertainty. Initially, influence diagrams were proposed as a front-end for decision trees [Howard and Matheson 1984]. Subsequently, Olmsted [1983] and Shachter [1986] developed methods for evaluating an influence diagram directly without converting it to a decision tree. These methods assume that all uncertain variables in the model are represented by discrete probability mass functions (PMF's). Several improvements to solution procedures for solving discrete influence diagrams have been proposed [see, e.g., Shenoy 1992, Shachter and Ndilikilikesha 1993, Jensen *et al.* 1994, Madsen and Jensen 1999, Lauritzen and Nilsson 2001, Madsen and Nilsson 2001].

Shachter and Kenley [1989] introduced Gaussian influence diagrams, which contain continuous variables with Gaussian distributions and a quadratic value function. In this framework, chance nodes have conditional linear Gaussian distributions, meaning each chance variable has a Gaussian distribution whose mean is a linear function of the variable's parents and whose variance is a constant. This framework does not allow discrete chance nodes; however, it does allow chance variables whose distributions are conditionally deterministic linear functions of their parents.

Poland and Shachter [1993] introduce mixture of Gaussians influence diagrams, which allow both discrete and continuous nodes where continuous variables are modeled as mixtures of Gaussians. In this framework, instantiating all discrete nodes reduces the model to a Gaussian influence diagram. The influence diagram must satisfy the condition that discrete chance nodes cannot have continuous parents. In this model, a quadratic value function is specified along with a utility function which represents risk-neutral behavior or a constant risk aversion. Poland [1993] proposes a procedure for solving such influence diagrams which uses discrete and Gaussian operations and reduces continuous chance variables before discrete chance variables. Madsen and Jensen [2003] describe a new procedure for exact evaluation of similar influence diagrams that contain an additively decomposing quadratic utility function.

Mixtures of truncated exponentials (MTE) potentials are suggested by Moral *et al.* [2001] and Rumí [2003] as an alternative to discretization for solving Bayesian networks with a mixture of discrete and continuous chance variables. In this paper, we propose MTE influence diagrams, which are influence diagrams in which probability distributions and utility functions are represented by MTE potentials. We solve MTE influence diagrams using the fusion algorithm proposed by Shenoy [1993] for the case where the joint utility function decomposes multiplicatively.

The remainder of this paper is organized as follows. Section 2 introduces notation and definitions used throughout the paper. Section 3 defines MTE potentials and presents a method for approximating joint utility functions with MTE utility potentials. Section 4 reviews the operations required for solving MTE influence diagrams. Section 5 contains an adaptation of Raiffa's [1968] Oil Wildcatter problem, which is represented and solved using an MTE influence diagram. Finally, section 6 summarizes and states some directions for future research. This paper is based on a larger, unpublished working paper [Cobb and Shenoy 2004].

## 2 NOTATION AND DEFINITIONS

This section contains notation and definitions used throughout the paper.

Variables will be denoted by capital letters, e.g. $A, B, C$. Sets of variables will be denoted by boldface capital letters, $\mathbf{Y}$ if all are discrete chance variables, $\mathbf{Z}$ if all are continuous chance variables, $\mathbf{D}$ if all are decision variables, or $\mathbf{X}$ if the components are a mixture of discrete chance, continuous chance, and decision variables. In this paper, all decision variables are assumed to be discrete. If $\mathbf{X}$ is a set of variables, $\mathbf{x}$ is a configuration of specific states of those variables. The discrete, continuous, or mixed state space of $\mathbf{X}$ is denoted by $\Omega_{\mathbf{X}}$.

MTE probability potentials and discrete probability potentials are denoted by lower-case greek letters, e.g. $\alpha$, $\beta$, $\gamma$. Subscripts are used for fragments of MTE potentials or conditional probability tables when different parameters or values are required for each configuration of a variable's discrete parents, e.g. $\alpha_1$, $\beta_2$, $\gamma_3$. Subscripts are also used for discrete probabilities of elements of the state space, e.g. $\delta_0 = P(D = 0)$.

MTE utility potentials are denoted by $u_i$, where the subscript $i$ indexes both the initial MTE utility potential(s) specified in the influence diagram and subsequent MTE utility potentials created during the solution procedure.

In graphical representations, decision variables are represented by rectangular nodes, discrete chance variables are represented by single-border ovals, continuous chance variables are represented by double-border ovals, and utility functions are represented by diamonds.

## 3 MIXTURES OF TRUNCATED EXPONENTIALS

### 3.1 MTE POTENTIALS

A mixture of truncated exponentials (MTE) potential in an influence diagram has the following definition, which is a modification of the original definition proposed by Moral *et al.* [2001] and Rumí [2003].

*MTE potential.* Let $\mathbf{X}$ be a mixed $n$-dimensional variable. Let $\mathbf{Y} = (Y_1, \ldots, Y_f)$, $\mathbf{Z} = (Z_1, \ldots, Z_c)$, and $\mathbf{D} = (D_1, \ldots, D_g)$ be the discrete chance, continuous chance, and decision variable parts of $\mathbf{X}$, respectively, with $c + f + g = n$. A function $\phi : \Omega_{\mathbf{X}} \mapsto \mathbb{R}^+$ is an MTE potential if one of the next two conditions holds:

1. The potential $\phi$ can be written as

$$\phi(\mathbf{x}) = \phi(\mathbf{y}, \mathbf{z}, \mathbf{d}) = a_0 + \sum_{i=1}^{m} a_i \exp\Bigg\{ \sum_{j=1}^{f} b_i^{(j)} y_j + \sum_{k=1}^{c} b_i^{(f+k)} z_k + \sum_{\ell=1}^{g} b_i^{(c+f+\ell)} d_\ell \Bigg\} \quad (1)$$

for all $\mathbf{x} \in \Omega_{\mathbf{X}}$, where $a_i, i = 0, \ldots, m$ and $b_i^{(j)}$, $i = 1, \ldots, m$, $j = 1, \ldots, n$ are real numbers.

2. There is a partition $\Omega_1, \ldots, \Omega_k$ of $\Omega_{\mathbf{X}}$ verifying that the domain of continuous chance variables, $\Omega_{\mathbf{Z}}$, is divided into hypercubes, the domain of the discrete chance and decision variables, $\Omega_{\mathbf{Y} \cup \mathbf{D}}$, is divided into arbitrary sets, and such that $\phi$ is defined as

$$\phi(\mathbf{x}) = \phi_i(\mathbf{x}) \qquad \text{if } \mathbf{x} \in \Omega_i, \quad (2)$$

where each $\phi_i, i = 1, ..., k$ can be written in the form of equation (1) (i.e. each $\phi_i$ is an MTE potential on $\Omega_i$).

In the definition above, $k$ is the number of *pieces*, and $m$ is the number of exponential *terms* in each piece of the MTE potential. Moral *et al.* [2002] proposes an iterative algorithm based on least squares approximation to estimate MTE potentials from data. Moral *et al.* [2003] describes a method to approximate conditional MTE potentials using a mixed tree structure. Cobb and Shenoy [2003] presents MTE approximations to the normal probability density function (PDF). Cobb *et al.* [2003] describes a nonlinear optimization procedure used to fit MTE parameters for approximations to standard PDF's, including the uniform, exponential, gamma, beta, and lognormal distributions. A 2-piece, 3-term un-normalized MTE potential which approximates the normal PDF is

$$\psi'(x) = \begin{cases} \sigma^{-1}(-0.0105643 \\ \quad +197.0557202 \exp\{2.2568434(\frac{x-\mu}{\sigma})\} \\ \quad -461.4392506 \exp\{2.3434117(\frac{x-\mu}{\sigma})\} \\ \quad +264.7930371 \exp\{2.4043270(\frac{x-\mu}{\sigma})\}) \\ \qquad \text{if } \mu - 3\sigma \leq x < \mu \\ \sigma^{-1}(-0.0105643 \\ \quad +197.0557202 \exp\{-2.2568434(\frac{x-\mu}{\sigma})\} \\ \quad -461.4392506 \exp\{-2.3434117(\frac{x-\mu}{\sigma})\} \\ \quad +264.7930371 \exp\{-2.4043270(\frac{x-\mu}{\sigma})\}) \\ \qquad \text{if } \mu \leq z \leq \mu + 3\sigma \\ 0 \qquad \text{elsewhere.} \end{cases} \quad (3)$$

Figure 1 shows a graph of the 2-piece, 3-term MTE approximation overlayed on the actual normal PDF for the case where $\mu = 0$ and $\sigma^2 = 1$ over the domain $[-3, 3]$. A normalized version of the 2-piece, 3-term MTE approximation to the normal PDF is

$$\psi(x) = (1/0.9973) \cdot \psi'(x). \quad (4)$$

Properties of this approximation are described in [Cobb and Shenoy 2003].

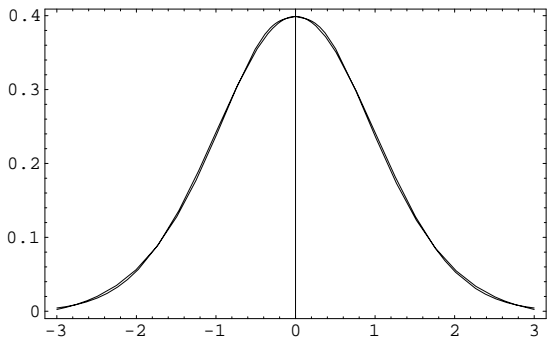

Figure 1: A 2-piece, 3-term MTE Approximation Overlayed on the Standard Normal Distribution

### 3.2 MTE PROBABILITY DENSITIES

*MTE probability densities.* Suppose $\phi'$ is an input MTE potential for $\mathbf{X} = \mathbf{Y} \cup \mathbf{Z} \cup \mathbf{D}$ representing a PDF for $Z \in \mathbf{Z}$ given its parents $\mathbf{X} \setminus \{Z\}$. If we can verify that

$$K(\mathbf{x}) = \int_{\Omega_Z} \phi'(\mathbf{x}, z) \, dz = 1 \,, \quad (5)$$

for all $\mathbf{x} \in \Omega_{\mathbf{X} \setminus Z}$, we state that $\phi'$ is an MTE density for $Z$. If $K(\mathbf{x}) \neq 1$ for any $\mathbf{x} \in \Omega_{\mathbf{X} \setminus Z}$, $\phi'$ can be normalized to form an MTE density by calculating $\phi = K(\mathbf{x})^{-1} \cdot \phi'$ for all $\mathbf{x} \in \Omega_{\mathbf{X} \setminus Z}$. We assume that all input MTE potentials which represent probability potentials in an MTE influence diagram are normalized to be MTE probability densities prior to the solution phase.

### 3.3 ESTIMATING MTE UTILITY POTENTIALS

In this paper, we consider problems with one joint utility function or a joint utility function which factors multiplicatively. The utility function(s) can be of any form as long as one can approximate them using MTE potentials. For instance, consider the case of a polynomial utility function. To create an MTE potential which approximates the joint utility function, we create an MTE approximation $\phi(x_i) = a_0 + a_1 \exp\{a_2 x_i\}$ for each variable $X_i$ in the polynomial utility function, then re-combine the MTE approximations and constants into a joint MTE utility potential. The result is an MTE utility potential because the class of MTE potentials is closed under addition and multiplication [Moral *et al.* 2001].

*Example 1.* Consider the joint utility function $u(x, y, z) = 3x^2 y + 4z^2 + 3xz^2 + 3y^2$.

This utility function can be decomposed into constants and the following linear functions: $g_1(x) = x, g_2(y) = y, g_3(z) = z$. Thus, $u(x, y, z)$ can be restated as

$$u(x, y, z) = 3 \cdot [g_1(x)]^2 \cdot g_2(y) + 4 \cdot [g_3(z)]^2 \\ + 3 \cdot g_1(x) \cdot [g_3(z)]^2 + 3 \cdot [g_2(y)]^2 \,.$$

The functions $g_1(x)$, $g_2(y)$, and $g_3(z)$ are approximated by MTE approximations $\phi_1(x)$, $\phi_2(y)$, and $\phi_3(z)$.

To create the MTE approximations $\phi(x_i)$ we use unconstrained, non-linear optimization and solve

$$\operatorname*{argmin}_{a_0, a_1, a_2} \sum_{j=0}^{n} (x_{ij} - \phi(x_{ij}))^2$$

where $x_{ij}, j = 0, ..., n$ is a set of points obtained by evenly dividing the domain of $X_i$. We then replace each $X_i$ in the utility function with its MTE approximation $\phi(x_i)$ and simplify the function accordingly.

A more detailed explanation of this procedure can be found in [Cobb and Shenoy 2004].

### 3.4 MTE INFLUENCE DIAGRAMS

An MTE influence diagram is an influence diagram in which all probability distributions are MTE probability densities as in (1) which satisfy the normalization condition in (5) and the joint utility function or the multiplicative factors of the joint utility function are MTE potential(s) as in (1). We assume decision nodes have discrete state spaces so that we stay in the class of MTE potentials during the solution process and also avoid optimization problems associated with continuous state spaces.

## 4 OPERATIONS ON MTE INFLUENCE DIAGRAMS

This section will describe the operations required to solve MTE influence diagrams.

### 4.1 COMBINATION

Combination of MTE potentials is pointwise multiplication. Let $\phi_1$ and $\phi_2$ be MTE potentials for $\mathbf{X}_1 = \mathbf{Y}_1 \cup \mathbf{Z}_1 \cup \mathbf{D}_1$ and $\mathbf{X}_2 = \mathbf{Y}_2 \cup \mathbf{Z}_2 \cup \mathbf{D}_2$. The combination of $\phi_1$ and $\phi_2$ is a new MTE potential for $\mathbf{X} = \mathbf{X}_1 \cup \mathbf{X}_2$ defined as follows

$$\phi(\mathbf{x}) = \phi_1(\mathbf{x}^{\downarrow \Omega_{\mathbf{X}_1}}) \cdot \phi_2(\mathbf{x}^{\downarrow \Omega_{\mathbf{X}_2}}) \qquad (6)$$

for all $\mathbf{x} \in \Omega_{\mathbf{X}}$.

Combination of two MTE probability densities results in an MTE probability density. Combination of an MTE probability density and an MTE utility potential results in an MTE utility potential. Combination of two MTE utility potentials results in an MTE utility potential. Combination of an MTE potential consisting of $k_1$ pieces with an MTE potential consisting $k_2$ pieces results in an MTE potential consisting of at most $k_1 \cdot k_2$ pieces. If the domains of the potentials do no overlap in all pieces, the resulting MTE potential may have less than $k_1 \cdot k_2$ pieces.

### 4.2 MARGINALIZATION

#### 4.2.1 Chance Variables

Marginalization of chance variables in an MTE influence diagram corresponds to summing over discrete chance variables and integrating over continuous chance variables. Let $\phi$ be an MTE potential for $\mathbf{X} = \mathbf{Y} \cup \mathbf{Z} \cup \mathbf{D}$. The marginal of $\phi$ for a set of variables $\mathbf{X}' = \mathbf{Y}' \cup \mathbf{Z}' \cup \mathbf{D} \subseteq \mathbf{X}$ is an MTE potential computed as

$$\phi^{\downarrow \mathbf{X}'}(\mathbf{y}', \mathbf{z}', \mathbf{d}) = \sum_{\mathbf{y} \in \Omega_{\mathbf{Y} \setminus \mathbf{Y}'}} \left( \int_{\Omega_{\mathbf{Z} \setminus \mathbf{z}'}} \phi(\mathbf{y}, \mathbf{z}, \mathbf{d}) \, d\mathbf{z}'' \right) \qquad (7)$$

where $\mathbf{z} = (\mathbf{z}', \mathbf{z}'')$, and $(\mathbf{y}', \mathbf{z}', \mathbf{d}) \in \Omega_{\mathbf{X}'}$.

Although we show the continuous variables being marginalized before the discrete variables in (7), the variables can be marginalized in any sequence, resulting in the same final MTE potential.

#### 4.2.2 Decision Variables

Marginalization with respect to a decision variable is only defined for MTE utility potentials. Let $u$ be an MTE utility potential for $\mathbf{X} = \mathbf{Y} \cup \mathbf{Z} \cup \mathbf{D}$, where $D \in \mathbf{D}$. The marginal of $u$ for a set of variables $\mathbf{X} - \{D\}$ is an MTE utility potential computed as

$$u^{\downarrow (\mathbf{X} - \{D\})}(\mathbf{y}, \mathbf{z}, \mathbf{d}') = \max_{d \in \Omega_D} u(\mathbf{y}, \mathbf{z}, \mathbf{d}) \qquad (8)$$

for all $(\mathbf{y}, \mathbf{z}, \mathbf{d}') \in \Omega_{\mathbf{X} - \{D\}}$ where $\mathbf{d} = (\mathbf{d}', d)$.

In order to use the fusion algorithm to solve MTE influence diagrams, marginalization of decision variables must result in an MTE potential. The following theorem ensures this result.

> **Theorem 1.** Let $u_1$ be an MTE utility potential for $\mathbf{X} = \mathbf{Y} \cup \mathbf{Z} \cup \mathbf{D}$, where $D \in \mathbf{D}$. If $u_1(\mathbf{y}, \mathbf{z}, \mathbf{d}', d = 1), ..., u_1(\mathbf{y}, \mathbf{z}, \mathbf{d}', d = n)$ are MTE utility potential fragments defined over the same domain, $\Omega_{\mathbf{X}}$, then $u_1^{\downarrow (\mathbf{X} - \{D\})} = \mathrm{Max}\{u_1(\mathbf{y}, \mathbf{z}, \mathbf{d}', d = 1), ..., u_1(\mathbf{y}, \mathbf{z}, \mathbf{d}', d = n)\}$ can be represented as an MTE utility potential whose components are equal to one of the fragments $u_1(\mathbf{y}, \mathbf{z}, \mathbf{d}', d = 1), ..., u_1(\mathbf{y}, \mathbf{z}, \mathbf{d}', d = n)$ in each region of a hypercube of $\Omega_{\mathbf{Z}}$, where $\mathbf{Z}$ are the continuous chance variables in $\mathbf{X}$.

A proof is given in [Cobb and Shenoy 2004]. Space constraints preclude reproducing the proof here.

### 4.3 Fusion Algorithm

A fusion algorithm for solving influence diagrams is described in Shenoy [1993]. The fusion algorithm involves deleting variables from the network in a sequence which respects the information constraints (represented by arcs pointing to decision variables in influence diagrams) in the problem. This condition ensures that unobserved chance variables are deleted before decision variables. The fusion method applies to problems where there is only one utility function (or a joint utility function which factors multiplicatively into several utility potentials) and uses only the operations of combination and marginalization as described above. Moral *et al.* [2001] shows that the class of MTE potentials is closed under marginalization (of chance variables) and combination. Theorem 1 states that the class of MTE potentials is closed under marginalization of discrete decision variables. Thus, MTE influence diagrams can be solved using the fusion algorithm, since only combinations and marginalizations are performed.

## 5 EXAMPLE

This example is an adaptation of the Oil Wildcatter problem from Raiffa [1968]. We model some variables as continuous uncertainties. Explicit representations of some MTE potentials are omitted, but can be found in [Cobb and Shenoy 2004].

An oil wildcatter must decide whether to drill ($D = 1$) or not drill ($D = 0$). He is uncertain whether the hole is dry ($O = 0$), wet ($O = 1$), or soaking ($O = 2$). The oil volume ($V$) extracted depends on the state of oil ($O$). If the hole is dry ($O = 0$), no oil is extracted. If $O = 1$, the amount of oil extracted follows a normal distribution with a mean of 6 thousand barrels and a standard deviation of 1 thousand barrels, i.e. $\mathcal{L}(V \mid O = 1) \sim N(6, 1^2)$. If $O = 2$, the amount of oil extracted follows a normal distribution with a mean of 13.5 thousand barrels and a standard deviation of 2 thousand barrels, i.e. $\mathcal{L}(V \mid O = 2) \sim N(13.5, 2^2)$. The cost of drilling ($C$) is normally distributed with a mean of 70 thousand dollars and a standard deviation of 10 thousand dollars, i.e. $\mathcal{L}(C) \sim N(70, 10^2)$. The log of oil prices ($P$) follows a normal distribution with a mean of $2.75 and a standard deviation of $0.7071, i.e. $\mathcal{L}(P) \sim LN(2.75, 0.7071^2)$.

The wildcatter assumes potential $\theta$ for $O$ as follows:

$\theta_0 = P(O = 0) = 0.500$
$\theta_1 = P(O = 1) = 0.300$
$\theta_2 = P(O = 2) = 0.200.$

At a cost of 10 thousand dollars, the wildcatter can conduct a seismic test which will help determine the geological structure at the site. The test results ($R$) will disclose whether the structure under the site has no structure ($R = 0$) (bad), open structure ($R = 1$) (so-so), or closed structure ($R = 2$) (very hope-ful). Experts have provided Table 1 which shows the probabilities of test results ($R$) conditional on the state of oil ($O$) and test ($T$) (which we will refer to as potential $\delta$ for $\{R, O, T\}$).

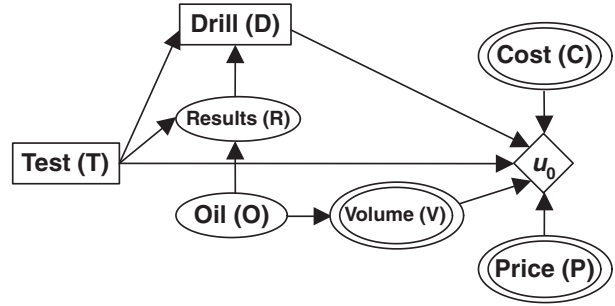

Figure 2: A Hybrid Influence Diagram Representation of the Oil Wildcatter Problem with Continuous Uncertainties

Table 1: Probabilities of Seismic Test Results Conditional on the Amount of Oil and Test

| Amount of Oil ($O$) | Structure Revealed by Seismic Test Results ($R$) | | | |
|---|---|---|---|---|
| | No Str. | Open Str. | Closed Str. | No Res. |
| $P(R \mid O, T = 1)$ | $R = 0$ | $R = 1$ | $R = 2$ | $NR$ |
| Dry ($O = 0$) | 0.60 | 0.30 | 0.10 | 0 |
| Wet ($O = 1$) | 0.30 | 0.40 | 0.30 | 0 |
| Soaking ($O = 2$) | 0.10 | 0.40 | 0.50 | 0 |
| $P(R \mid O, T = 0)$ | | | | |
| Dry ($O = 0$) | 0 | 0 | 0 | 1 |
| Wet ($O = 1$) | 0 | 0 | 0 | 1 |
| Soaking ($O = 2$) | 0 | 0 | 0 | 1 |

Figure 2 shows a hybrid influence diagram representation of the Oil Wildcatter problem with some discrete and some continuous chance variables.

### 5.1 Representation

The single utility function in the problem has domain $\{C, P, V, D, T\}$ and can be stated (in $000) as

$u_0(v, p, c, D = 1, T = 1) = v \cdot p - c - 10$
$u_0(v, p, c, D = 1, T = 0) = v \cdot p - c$
$u_0(v, p, c, D = 0, T = 1) = -10$
$u_0(v, p, c, D = 0, T = 0) = 0.$

The utility function $u_0$ can be approximated by an

MTE potential $u_1$ by using the method described in Section 3.3. The resulting MTE utility potential is

$u_1(v, p, c, D = 1, T = 1) =$
 $600,462,529.9767685$
 $+24,504.975886 \exp\{-0.00004109695c\}$
 $-600,488,161.2450081 \exp\{0.00004069868p\}$
 $+600,488,190.477144 \exp\{0.00004069868p$
  $+0.00004078953v\}$
 $-600,487,073.8393291 \exp\{0.00004078953v\}$

$u_1(v, p, c, D = 1, T = 0) =$
 $u_1(v, p, c, D = 1, T = 1) - 10$

$u_1(v, p, c, D = 0, T = 1) = -10$
$u_1(v, p, c, D = 0, T = 0) = 0 \,.$

Normally distributed chance variables are modeled using the 2-piece MTE approximation to the normal PDF given in (4). The potential fragments $\nu_1(v, O = 1)$ and $\nu_2(v, O = 2)$, which constitute the potential $\nu$ for $\{V, O\}$, are displayed graphically in Figure 3. The numerical descriptions of these potential fragments, as well as the MTE potential $\vartheta$ for $C$, are omitted.

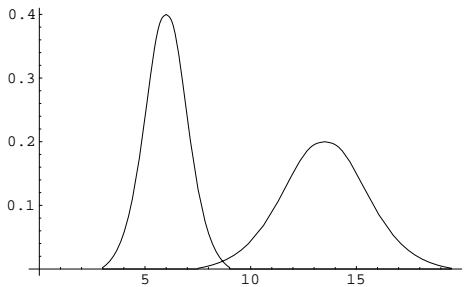

Figure 3: MTE Approximations of the PDF's for $V$ given $O$

An MTE approximation of the lognormal PDF $\rho$ for $P$ is constructed using the procedure in Cobb et al. [2003]. This MTE potential is as follows:

$\rho(p) = P(P) =$

$\begin{cases}
-0.024921 \\
+0.186834 \exp\{0.249714(p - 9.44687)\} \\
+0.101347 \exp\{1.419659(p - 9.44687)\} \\
\quad \text{if } 1.86706 \leq p < 3.47531 \\
0.174804 \\
-0.062119 \exp\{-0.116729(p - 9.44687)\} \\
-0.066038 \exp\{0.116608(p - 9.44687)\} \\
\quad \text{if } 3.47531 \leq p < 9.44687 \\
0.049064 \\
+0.000000154912 \exp\{1.480552(p - 9.44687)\} \\
-0.002427 \exp\{0.287079(p - 9.44687)\} \\
\quad \text{if } 9.44687 \leq p < 15.57526 \\
-0.583002 \\
+0.057534 \exp\{-0.079477(p - 9.44687)\} \\
+0.584025 \exp\{-0.000015(p - 9.44687)\} \\
\quad \text{if } 15.57526 \leq p \leq 129.93107 \\
0 \quad \text{elsewhere.}
\end{cases}$

The potential $\rho$ is displayed graphically in Figure 4 overlayed on the actual $LN(2.75, 0.7071^2)$ distribution.

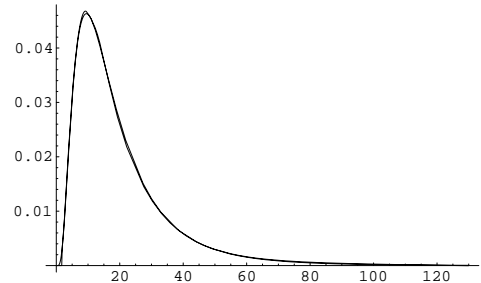

Figure 4: The MTE Approximation for the Distribution of $P$ Overlayed on the Actual $LN(2.75, 0.7071^2)$ Distribution

## 5.2 Solution

To calculate the optimal strategy and expected profit associated with that strategy, we use the fusion algorithm and delete the variables in the sequence $C, P, V, O, D, R, T$.

To remove $C$, we calculate $u_2 = (u_1 \otimes \vartheta)^{\downarrow\{P,V,D,T\}}$. To remove $P$, we calculate $u_3 = (u_2 \otimes \rho)^{\downarrow\{V,D,T\}}$. To remove $V$, we calculate $u_4 = (u_3 \otimes \nu)^{\downarrow\{O,D,T\}}$, which is described in Table 2.

The potentials remaining in the network after removal of $V$ are $u_4$ with domain $\{O, D, T\}$, $\delta$ with

Appeared in: M. Chickering and J. Halpern (eds.), Uncertainty in Artificial Intelligence (UAI-04), 2004, pp. 85--93, AUAI Press

Table 2: The Utility Function $u_4$ with Domain $\{O, D, T\}$ Resulting from the Removal of Variable $V$ ($000)

|  | Values of Drill ($D$) and Test ($T$) | | | |
|---|---|---|---|---|
| Amount of Oil ($O$) | $D=1$ $T=1$ | $D=1$ $T=0$ | $D=0$ $T=1$ | $D=0$ $T=0$ |
| Dry ($O=0$) | −82.75 | −72.75 | −10.00 | 0 |
| Wet ($O=1$) | 40.95 | 50.95 | −10.00 | 0 |
| Soak. ($O=2$) | 192.22 | 202.22 | −10.00 | 0 |

domain $\{R, O, T\}$ and $\theta$ with domain $\{O\}$. Thus, to remove $O$, we calculate $u_5 = (u_4 \otimes \theta \otimes \delta)^{\downarrow\{D,R,T\}}$, which is described in Table 3.

Table 3: The Utility Function $u_5$ with Domain $\{D, R, T\}$ Resulting from the Removal of Variable $O$ ($000)

|  | Values of Drill ($D$) and Test ($T$) | | | |
|---|---|---|---|---|
| Results of Test ($R$) | $D=1$ $T=1$ | $D=1$ $T=0$ | $D=0$ $T=1$ | $D=0$ $T=0$ |
| No Result | 0 | 19.35 | 0 | 0 |
| No Str. ($R=0$) | −17.30 | 0 | −4.10 | 0 |
| Open Str. ($R=1$) | 7.88 | 0 | −3.50 | 0 |
| Clsd. Str. ($R=2$) | 18.77 | 0 | −2.40 | 0 |

Removing $D$ involves simply maximizing the utility in Table 3 for each configuration of $\{R, T\}$. The resulting utility function $u_6$ is shown in Table 4. The optimal policy is drill ($D = 1$) if a test is performed and the results reveal open structure ($R = 1$) or closed structure ($R = 2$), not drill ($D = 0$) if a test is performed and the results reveal no structure ($R = 0$), and drill ($D = 1$) if no test is performed.

Summing the values in Table 4 over the possible values of $R$ gives $u_7(T = 1) = -4.10 + 7.88 + 18.77 = 22.55$ and $u_7(T = 0) = 19.35$. Thus, the optimal test decision is to test ($T = 1$), and the maximum expected profit is $22,550.

### 5.3 Continuous Test Results

Suppose that the seismic test in the Oil Wildcatter example yields a continuous reading ($R$) representing the location of the peak response, measured on the unit interval [0,1]. The PDF's for $R$ given $O$ and $T = 1$ are symmetric beta distributions as follows:

$$\pounds(R \mid O = 0, T = 1) \sim Beta(1, 1)$$

Table 4: The Utility Function $u_6$ with Domain $\{R, T\}$ Resulting from the Removal of Variable $D$ ($000), with Optimal Policies

| Results of Test ($R$) | Value of Test ($T$) | |
|---|---|---|
|  | $T=1$ | $T=0$ |
| No Result | 0 | 19.35 ($D$=1) |
| No Str. ($R=0$) | −4.10 ($D$=0) | 0 |
| Open Str. ($R=1$) | 7.88 ($D$=1) | 0 |
| Clsd. Str. ($R=2$) | 18.77 ($D$=1) | 0 |

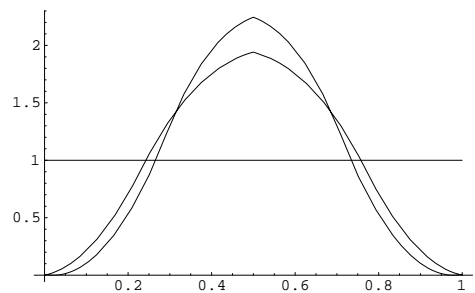

Figure 5: MTE Approximations of the Probability Density Functions for $R$ given $O$ and $T = 1$.

$$\pounds(R \mid O = 1, T = 1) \sim Beta(3.2, 3.2)$$
$$\pounds(R \mid O = 2, T = 1) \sim Beta(4.2, 4.2).$$

These distributions are approximated by MTE potential fragments using the procedure in Cobb et al. [2003]. The MTE potential fragments $\delta_0(r, O = 0, T = 1)$, $\delta_1(r, O = 1, T = 1)$ and $\delta_2(r, O = 2, T = 1)$ constitute the potential fragment $\delta$ for $\{R, O, T = 1\}$. These fragments are shown in Figure 5—$\delta_0$ is the flat distribution, $\delta_2$ is the most peaked distribution, and $\delta_1$ is in-between. An observation in the middle of the unit interval will favor $O = 2$, an observation near the extremes $R = 0$ or 1 will favor $O = 0$, and an observation around $R = 0.275$ or $0.725$ will favor $O = 1$.

The solution remains the same as in Section 5.2 through the removal of $V$. To remove $O$, we calculate $u_5 = (u_4 \otimes \theta \otimes \delta)^{\downarrow\{D,R,T\}}$. The utility potential fragments $u_5(r, D = 1, T = 1)$ and $u_5(r, D = 0, T = 1)$ are shown graphically in Figure 6. The other utility potential fragments constituting the utility potential $u_5$ are constants: $u_5(r, D = 1, T = 0) = 19.354$ and $u_5(r, D = 0, T = 0) = 0$.

For $T = 1$, removing $D$ involves finding $\text{Max}\{u_5(r, D = 1, T = 1), u_5(r, D = 0, T = 1)\}$ at each point in the domain of $R$. We can recover an MTE potential from this calculation by identifying points where the two utility potential fragments are equal, creating new regions using these points, then selecting either $u_5(r, D = 1, T = 1)$ or $u_5(r, D = 0, T = 1)$ within each of these regions[1]. In this case, we find $u_5(r, D = 1, T = 1) \approx u_5(r, D = 0, T = 1)$ at 0.212 and 0.788. This results in utility function $u_6(r, T = 1)$ which appears graphically in Figure 7. The optimal strategy is drill if the test result is in the interval $[0.212, 0.788]$ and not drill otherwise. For $T = 0$, removing $D$ involves simply selecting the value of $D$ which yields the highest utility; thus, we select $D = 1$ which gives $u_6(r, T = 0) = 19.354$.

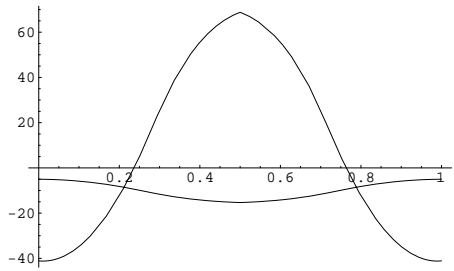

Figure 6: Utility fragments $u_5(r, D = 1, T = 1)$ (increasing on (0,0.5]) and $u_5(r, D = 0, T = 1)$.

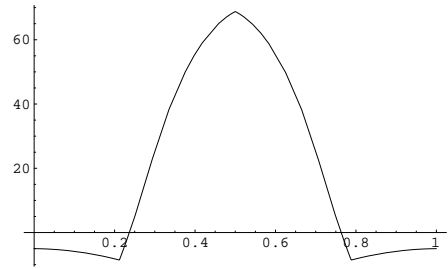

Figure 7: The utility potential fragment $u_6(r, T = 1)$.

Removing $R$ results in utility potential $u_7$, defined as follows:

$$u_7(T = 1) = \int_0^1 u_6(r, T = 1) \, dr = 19.802$$
$$u_7(T = 0) = 19.354\,.$$

Thus, the optimal decision is to test $(T = 1)$ and the maximum expected profit is $19,802.

## 6 CONCLUSIONS AND SUMMARY

We have described MTE influence diagrams and demonstrated a procedure for solving MTE influence diagrams with one joint utility function (or multiplicative factors of one joint utility function) when probability distributions are represented by MTE probability densities and utility functions are represented by MTE utility potentials. Any continuous PDF can be modeled by an MTE potential, so any continuous random variable can be represented in an MTE influence diagram. This includes, e.g. conditional linear Gaussian, gamma, beta, and lognormal distributions. The solution method presented places no restrictions on the arrangement of discrete and continuous chance variables in the influence diagram.

As described, MTE influence diagrams have some limitations. First, the numerical stability of the solution algorithm may be an issue in problems where the MTE approximations have very large and/or very small parameters. This needs further investigation. Second, MTE influence diagrams only allow for multiplicative factorization of the joint utility function. This is because solving an influence diagram with an additive factorization involves division of potentials, and the class of MTE potentials is not closed under division. Third, MTE influence diagrams only allow discrete decision variables. This is because Theorem 1, which states that the class of MTE potentials is closed under marginalization of decision variables, holds only for discrete decision variables. Also, marginalizing continuous decision variables from arbitrary MTE utility potentials is a complex optimization problem. This limitation needs further research.

### Acknowledgements

This paper has benefited from comments by Thomas D. Nielsen, Finn V. Jensen, and three anonymous reviewers. This research was partly funded by a graduate research assistantship provided to the first author from the Ronald G. Harper Professorship and by a contract from Sparta, Inc. to the second author.

---
[1] We use the bisection search method to perform this operation (for details, see [Cobb and Shenoy 2004]).